\documentclass[10pt,twocolumn,letterpaper]{article}

\usepackage{cvpr}
\usepackage{times}
\usepackage{epsfig}
\usepackage{graphicx}
\usepackage{amsmath}
\usepackage{amssymb}
\usepackage{float}
\usepackage{bbding}
\usepackage{authblk}

\usepackage[pagebackref=true,breaklinks=true,letterpaper=true,colorlinks,bookmarks=false]{hyperref}

\cvprfinalcopy 

\def\cvprPaperID{3966} 

\ifcvprfinal\fi
\begin{document}


\title{FS-Net: Fast Shape-based Network for Category-Level 6D Object Pose Estimation with Decoupled Rotation Mechanism}


\author{Wei Chen $^{1}$ \hspace{0.5cm} Xi Jia$^{1}$ \hspace{0.5cm} Hyung Jin Chang$^{1}$\hspace{0.5cm} Jinming Duan$^{1}$\hspace{0.5cm} Linlin Shen$^{2}$\hspace{0.5cm}  Ale\v{s} Leonardis$^{1}$\\
{\small{$^1$ University of Birmingham \quad $^2$Shenzhen University}}\\
 {\tt\small \Envelope wxc795@cs.bham.ac.uk}
}

\maketitle

\begin{abstract}
In this paper, we focus on category-level 6D pose and size estimation from monocular RGB-D image. Previous methods suffer from inefficient category-level pose feature extraction which leads to low accuracy and inference speed. To tackle this problem, we propose a fast shape-based network (FS-Net) with efficient category-level feature extraction for 6D pose estimation. First, we design an orientation aware autoencoder with 3D graph convolution for latent feature extraction. The learned latent feature is insensitive to point shift and object size thanks to the shift and scale-invariance properties of the 3D graph convolution. Then, to efficiently decode category-level rotation information from the latent feature, we propose a novel decoupled rotation mechanism that employs two decoders to complementarily access the rotation information. Meanwhile, we estimate translation and size by two residuals, which are the difference between the mean of object points and ground truth translation, and the difference between the mean size of the category and ground truth size, respectively. Finally, to increase the generalization ability of FS-Net, we propose an online box-cage based 3D deformation mechanism to augment the training data. Extensive experiments on two benchmark datasets show that the proposed method achieves state-of-the-art performance in both category- and instance-level 6D object pose estimation. Especially in category-level pose estimation, without extra synthetic data, our method outperforms existing methods by $6.3\%$ on the NOCS-REAL dataset \footnote{Paper code \url{https://github.com/DC1991/FS-Net}}.
\end{abstract}

\section{Introduction}
\begin{figure}[h!]
\begin{center}
\includegraphics[width=1.0\linewidth]{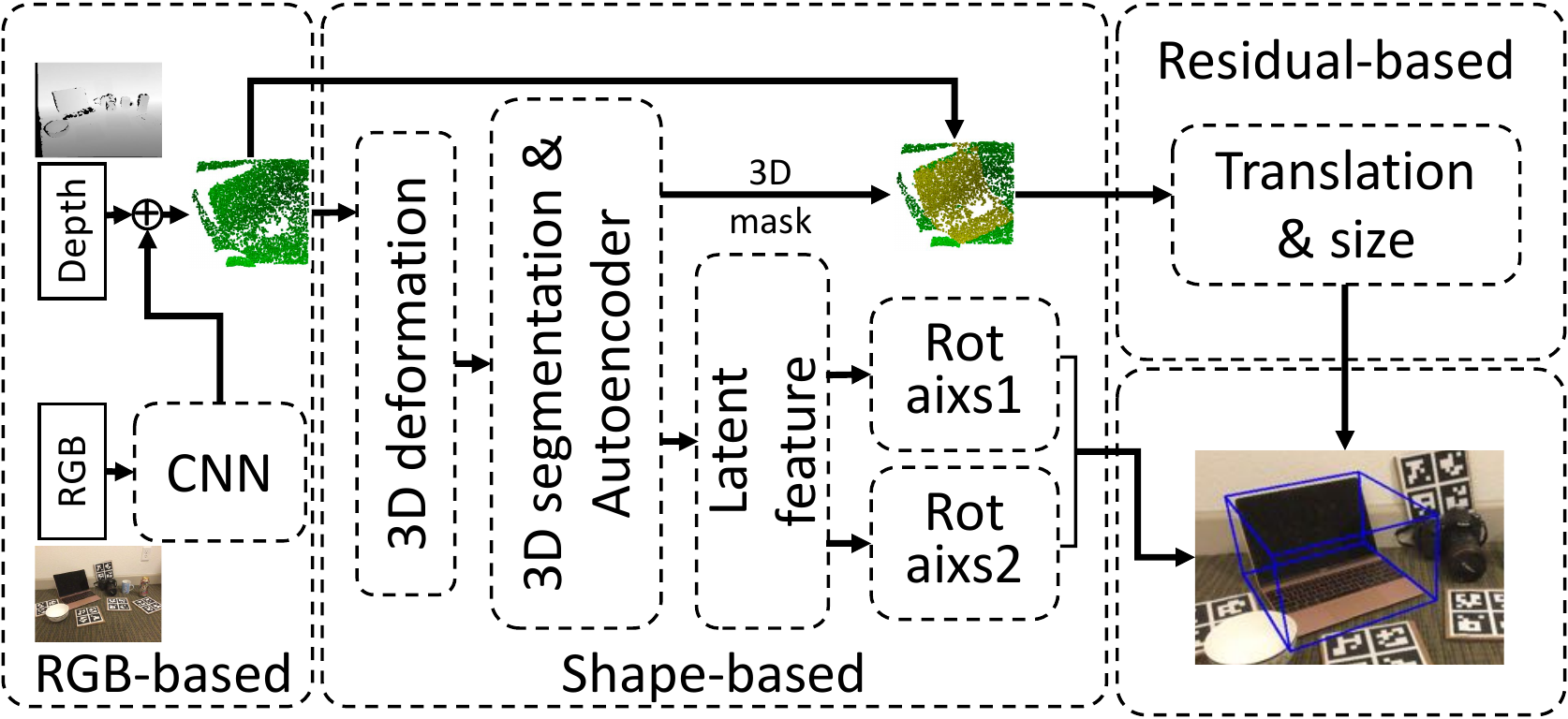}
\end{center}
   \caption{\textbf{Semantic illustration of FS-Net.} We use different networks for different tasks. The RGB-based network is used for 2D object detection, and the shape-based 3D graph convolution autoencoder is used for 3D segmentation and rotation estimation. The residual-based network is used for translation and size estimation with segmented points.}
\label{fig:short}
\vspace{-10pt}
\end{figure}
Estimating 6D object pose plays an essential role in many computer vision tasks such as augmented reality \cite{marchand2016pose, marder2016project}, virtual reality \cite{burdea2003virtual}, and smart robotic arm \cite{zhu2014single,tremblay2018deep}. 
For instance-level 6D pose estimation, in which training set and test set contain the same objects, huge progress has been made in recent years \cite{Xiang2017,Rad2017bb8,Oberweger2018,Li_2018_ECCV,he2020pvn3d}.

However, category-level 6D pose estimation remains challenging as the object shape and color are various in the same category.
Existing methods addressed this problem by mapping the different objects in the same category into a uniform model via RGB feature or RGB-D fusion feature. 
For example, Wang \textit{et al.} \cite{wang2019nocs} trained a modified Mask R-CNN \cite{he2017mask} to predict the normalized object coordinate space (NOCS) map of different objects based on RGB feature, and then computed the pose with observed depth and NOCS map by Umeyama algorithm \cite{umeyama1991least}. 
Chen \textit{et al.} \cite{chen2020cass} proposed to learn a canonical shape space (CASS) to tackle intra-class shape variations with RGB-D fusion feature \cite{wang2019densefusion}.
Tian \textit{et al.} \cite{tian2020shapeprior} trained a network to predict the NOCS map of different objects, with the uniform shape prior learned from a shape collection, and RGB-D fusion feature \cite{wang2019densefusion}. 

Although these methods achieved state-of-the-art performance, there are still two issues.
Firstly, the benefits of using RGB feature or RGB-D fusion feature for category-level pose estimation are still questionable.
In \cite{vlach2016we}, Vlach et al. showed that people focus more on shape than color when categorizing objects, as different objects in the same category have very different colors but stable shapes (shown in Figure \ref{fig:color_var}). Thereby the use of RGB feature for category-level pose estimation may lead to low performance due to huge color variation in the test scene. For this issue, to alleviate the color variation, we merely use the RGB feature for 2D detection, while using the shape feature learned with point cloud extracted from depth image for category-level pose estimation.

Secondly, learning a representative uniform shape requires a large amount of training data; therefore, the performance of these methods is not guaranteed with limited training examples. To overcome this issue, we propose a 3D graph convolution (3DGC) autoencoder \cite{lin2020convolution} to effectively learn the category-level pose feature via observed points reconstruction of different objects instead of uniform shape mapping. We further propose an online box-cage based 3D data augmentation mechanism to reduce the dependencies of labeled data.

In this paper, the newly proposed FS-Net consists of three parts: 2D detection, 3D segmentation \& rotation estimation, and translation \& size estimation. In 2D detection part, we use the YOLOv3 \cite{redmon2018yolov3} to detect the object bounding box for coarse object points obtainment \cite{Chen_2020_CVPR}. 
Then in the 3D segmentation \& rotation estimation part, we design a 3DGC autoencoder to perform segmentation and observed points reconstruction jointly. The autoencoder encodes orientation information in the latent feature. Then we propose the decoupled rotation mechanism that uses two decoders to decode the category-level rotation information.
For translation and size estimation, since they are all point coordinates related, we design a coordinate residual estimation network based on PointNet \cite{Qi_2017_CVPR} to estimate the translation residual and size residuals. 
To further increase the generalization ability of FS-Net, we use the proposed online 3D deformation for data augmentation.
%
To summarize, the main contributions of this paper are as follows:
\begin{itemize}

    \item We propose a fast shape-based network to estimate category-level 6D object size and pose. Due to the efficient category-level pose feature extraction, the framework runs at 20 FPS on a GTX 1080 Ti GPU.
    
    \item We propose a 3DGC autoencoder to reconstruct the observed points for latent orientation feature learning. Then we design a decoupled rotation mechanism to fully decode the orientation information. This decoupled mechanism allows us to naturally handle the circle symmetry object (in Section \ref{sec:rotde}).
    
    \item Based-on the shape similarity of intra-class objects, we propose a novel box-cage based 3D deformation mechanism to augment the training data. With this mechanism, the pose accuracy of FS-Net is improved by $7.7\%$.
\end{itemize}

\begin{figure*}[htp!]
\begin{center}
\includegraphics[width=0.9\linewidth]{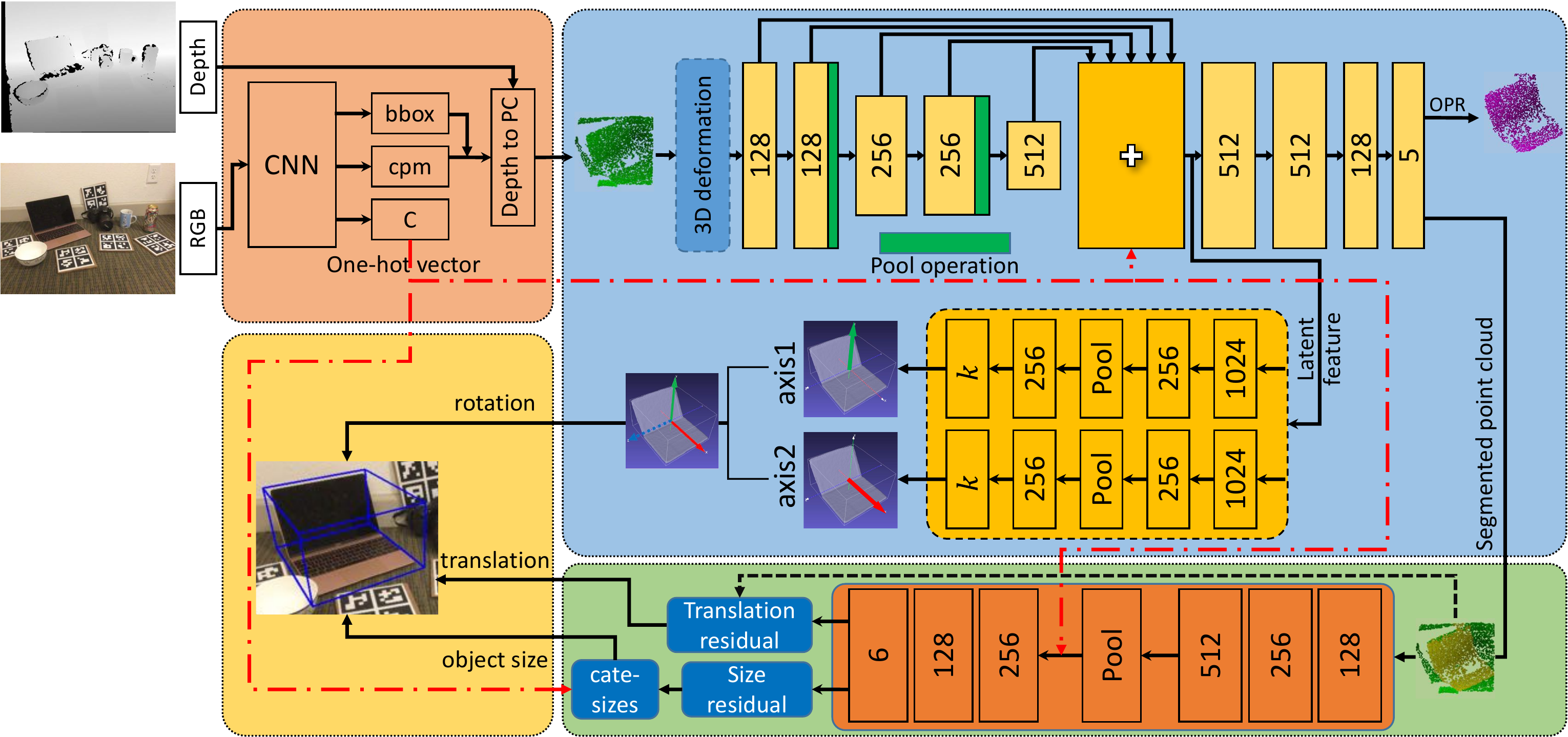}

\end{center}
\caption{\textbf{Architecture of FS-Net.} The input of FS-Net is an RGB-D image. For RGB channels, we use a 2D detector to detect the object 2D location, category label `C' (used as a one-hot feature for next tasks), and class probability map (cpm) (generate the 3D sphere center via maximum probability location and camera parameters). With this information and depth channel, the points in a compact 3D sphere are generated. Given the points in the 3D sphere, we first use the proposed 3D deformation mechanism for data augmentation. After that, we use a shape-based 3DGC autoencoder to perform observed points reconstruction (OPR), as well as point cloud segmentation, for orientation latent feature learning. Then we decode the rotation information into two perpendicular vectors from the latent feature. Finally, we use a residual estimation network to predict the translation and size residuals. `cate-sizes' denotes the pre-calculated average sizes of different categories, `$k$' is the rotation vector dimension, and the hollow `+' means feature concatenation.}
\label{fig:arch_whole}
\end{figure*}


\section{Related Works}
\subsection{Instance-Level Pose Estimation}
In instance-level pose estimation, a known 3D object model is usually available for training and testing. Based on the 3D model, instance-level pose estimation can be roughly divided into three types: template matching based, correspondences-based, and voting-based methods. Template matching methods \cite{hinterstoisser2012gradient, Rad2018, Oberweger2018} aligned the template to the observed image or depth map via hand-crafted or deep learning feature descriptors. As they need the 3D object model to generate the template pool, their applications in category-level 6D pose estimation are limited. Correspondences-based methods trained their model to establish 2D-3D correspondences \cite{Rad2017bb8,Rad2018,peng2018pvnet} or 3D-3D correspondences \cite{Chen_2020_CVPR, Chen_2020_WACV}. Then they solved perspective-n-point and SVD problem with 2D-3D and 3D-3D correspondences \cite{kabsch1976solution}, respectively. Some methods \cite{Chen_2020_WACV, Brachmann2016} also used these correspondences to generate voting candidates, and then used RANSAC \cite{fischler1981random} algorithm for selecting the best candidate. However, the generation of canonical 3D keypoints is based on the known 3D object model that is not available when predicting the category-level pose.

\subsection{Category-Level Pose Estimation}
Compared to instance-level, the major challenge of category-level pose estimation is the intra-class object variation, including shape and color variation. To handle the object variation problem, \cite{wang2019nocs} proposed to map the different objects in the same category to a NOCS map. Then they used semantic segmentation to access the observed points cloud with known camera parameters. The 6D pose and size are calculated by the Umeyama algorithm \cite{umeyama1991least} with the NOCS map and the observed points. Shape-Prior \cite{tian2020shapeprior} adopted similar method with \cite{wang2019nocs}, but both extra shape prior knowledge and dense-fusion feature \cite{wang2019densefusion}, instead of RGB feature, are used. CASS \cite{chen2020cass} estimated the 6D pose via the learning of a canonical shape space with dense-fusion feature \cite{wang2019densefusion}. Since the RGB feature is sensitive to color variation, the performance of their methods in category-level pose estimation is limited. In contrast, our method is shape feature-based which is robust for this task.

\subsection{3D Data Augmentation}
In 3D object detection tasks \cite{Chen_2020_CVPR,Qi_2018_CVPR,shi2020pv, Chen_2020_WACV}, online data augmentation techniques such as translation, random flipping, shifting, scaling, and rotation are applied to original point clouds for training data augmentation. However, these operations cannot change the shape property of the object. Simply adopting these operations on point clouds is not able to handle the shape variation problem in the 3D task. To address this, \cite{choi2020part} proposed part-aware augmentation which operates on the semantic parts of the 3D object with five manipulations: dropout, swap, mix, sparing, and noise injection. However, how to decide the semantic parts are ambiguous. In contrast, we propose a box-cage based 3D data augmentation mechanism which can generate the various shape variants (shown in Figure \ref{fig:3d_defor}) and avoid semantic parts decision procedure.

\section{Proposed Method}

In this section, we describe the detailed architecture of FS-Net shown in Figure \ref{fig:arch_whole}. 
Firstly, we use the YOLOv3 to detect the object location with RGB input. Secondly, we use 3DGC autoencoder to perform 3D segmentation and observed points reconstruction, the latent feature can learn orientation information through the process.
Then we propose a novel decoupled rotation mechanism for decoding orientation information. Thirdly, we use PointNet \cite{Qi_2017_CVPR} to estimate the translation and object size. Finally, to increase the generalization ability of FS-Net, we propose the box-cage based 3D deformation mechanism.

\begin{figure}[t!]
\begin{center}
\includegraphics[width=0.8\linewidth]{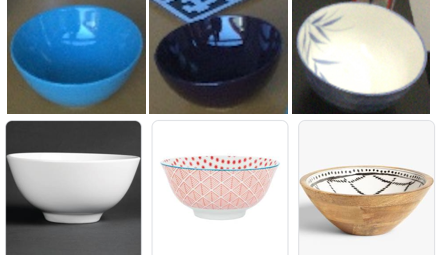}
\end{center}
   \caption{\textbf{Stable shape and various color.} Top row: three bowl instances randomly chosen from the NOCS-REAL dataset. Bottom row: three bowl instances randomly cropped from the internet image search results (using the keyword `bowl'). The color is varied, while the shape is relatively stable.}
\label{fig:color_var}
\vspace{5pt}
\end{figure}

\subsection{Object Detection}
\label{sec:obj_det}

Following \cite{Chen_2020_CVPR}, we train a YOLOv3 \cite{redmon2018yolov3} to fast detect the object bounding box in RGB images, and output class (category) labels. Then we adopt the 3D sphere to locate the point cloud of the target object quickly. 
With these techniques, the 2D detection part provides a compact 3D learning space for the following tasks. 
Different from other category-level 6D object pose estimation methods that need semantic segmentation masks, we only need object bounding boxes. Since object detection is faster than semantic segmentation \cite{redmon2018yolov3, he2017mask}, the detection speed of our method is faster than previous methods.

\subsection{Shape-Based Network}


The output points of object detection contain both object and background points.
To access the points that belong to the target object and calculate the rotation of the object, we need a network that performs two tasks: 3D segmentation and rotation estimation. 

Although there are many network architectures that directly process point cloud \cite{Qi_2017_CVPR,qi2017pointnetplusplus,zhou2018voxelnet}, most of the architectures calculate on point coordinates, which means their networks are sensitive to point clouds shift and size variation \cite{lin2020convolution}. This decreases the pose estimation accuracy.

To tackle the point clouds shift, Frustum-PointNet \cite{Qi_2018_CVPR} and G2L-Net \cite{Chen_2020_CVPR} employed the estimated translation to align the segmented point clouds to local coordinate space. However, their methods cannot handle the intra-class size variation.

To solve the point clouds shift and size variation problem, in this paper, we propose a 3DGC autoencoder to extract the point cloud shape feature for segmentation and rotation estimation. 3DGC is designed for point cloud classification and object part segmentation; our work shows that 3DGC can also be used for category-level 6D pose estimation task.

\subsubsection{3D Graph Convolution}
3DGC kernel consists of $m$ unit vectors. The $m$ kernel vectors are applied to the $n$ vectors generated by the center point with its $n$-nearest neighbors. Then, the convolution value is the sum of cosine similarity between kernel vectors and the $n$-nearest vectors. In a 2D convolution network, the trained network learned a weighted kernel, which has a higher response with a matched RGB value, while the 3DGC network learned the orientations of the $m$ vectors in the kernel. The weighted 3DGC kernel has a higher response with a matched 3D pattern which is defined by the center point with its $n$-nearest neighbors. For more details, please refer to \cite{lin2020convolution}.

\subsubsection{Rotation-Aware Autoencoder}
Based on the 3DGC, we design an autoencoder for the estimation of category-level object rotation. To extract the latent rotation feature, we train the autoencoder to reconstruct the observed points transformed from the observed depth map of the object. There are several advantages to this strategy: 1) the reconstruction of observed points is view-based and symmetry invariant \cite{sundermeyer2020multi,sundermeyer2018implicit}, 2) the reconstruction of observed points is easier than that of a complete object model (shown in Table \ref{tab:recon}), and 3) more representative orientation feature can be learned (shown in Table \ref{tab:abla}).

In \cite{sundermeyer2020multi,sundermeyer2018implicit}, the authors also reconstructed the input images to observed views. However, the input and output of their models are 2D images that are different to our 3D point cloud input and output. Furthermore, our network architecture is also different from theirs.

We utilize Chamfer Distance to train the autoencoder, the reconstruction loss function $\mathcal{L}_{rec}$ is defined as
\begin{equation}
\mathcal{L}_{rec} =
\sum_{x_i \in M_{c}} \min _{\hat{x}_i \in \hat{M}_{c}}\|x_i-\hat{x}_i\|_{2}^{2}
 +\sum_{\hat{x}_i \in \hat{M}_{c}} \min _{x_i \in M_{c}}\|x_i-\hat{x}_i\|_{2}^{2},
\end{equation}
where $M_c$ and $\hat{M}_c$ denote the ground truth point cloud and reconstructed point cloud, respectively. $x_i$ and $\hat{x}_i$ are the points in $M_c$ and $\hat{M}_c$. With the help of 3D segmentation mask, we only use the features extracted from the observed object points for reconstruction.

After the network convergence, the encoder learned the rotation-aware latent feature. 
Since the 3DGC is scale and shift invariant, the observed points reconstruction enforces the autoencoder to learn the scale and shift invariant orientation feature under corresponding rotation.
In the next subsection, we will describe how we decode rotation information from this latent feature.

\subsection{Decoupled Rotation Estimation}
\label{sec:rotde}
\begin{figure}[htp!]
\begin{center}
\includegraphics[width=0.8\linewidth, height=3.1cm]{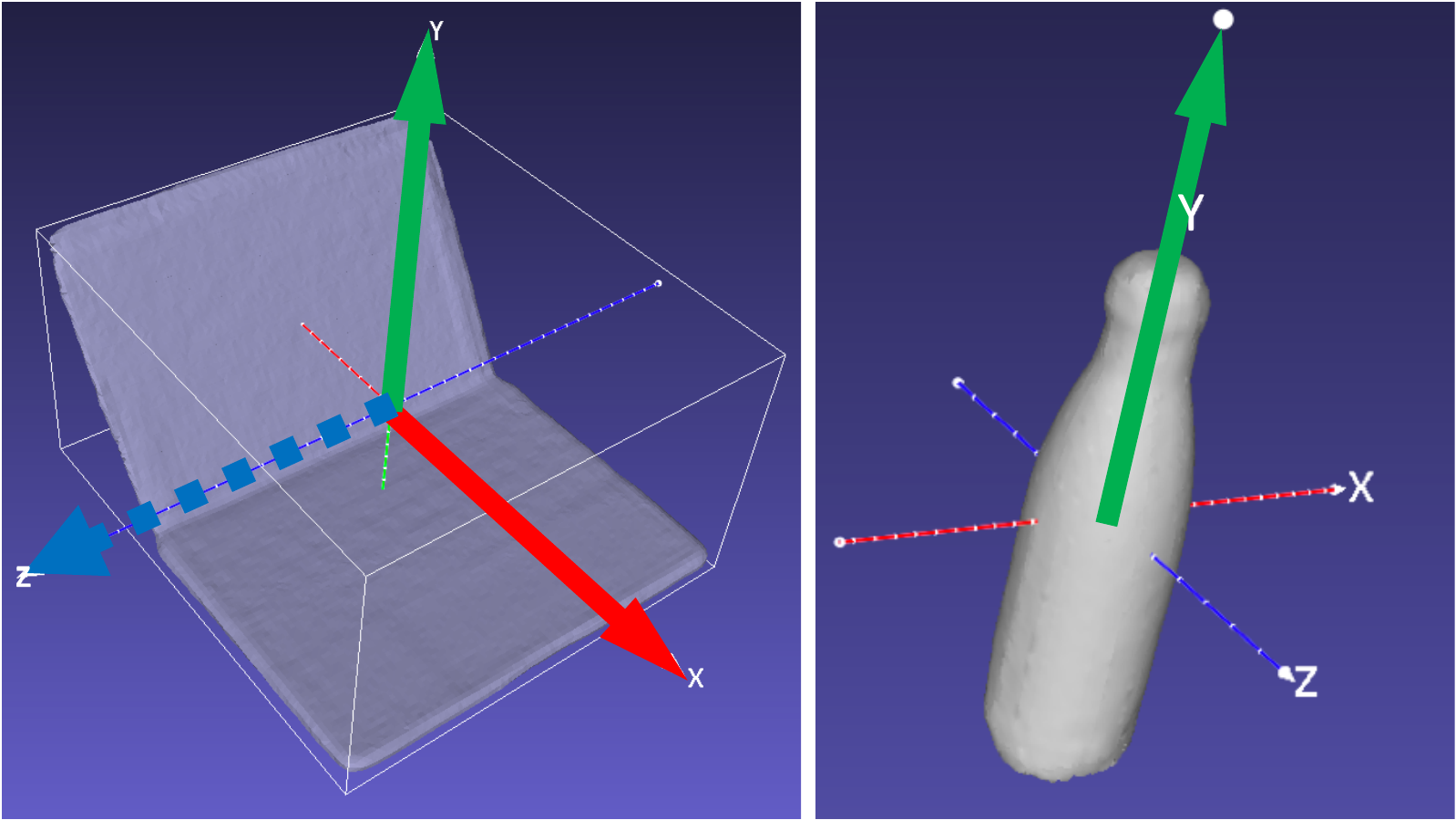}
\end{center}
   \caption{\textbf{Rotation represented by vectors.} Left: The object rotation can be represented by two perpendicular vectors (green vector and red vector); Right: For circular symmetry object like the bottle, only the green vector matters.}
\label{fig:axis_0}
\end{figure}

Given the latent feature which contains rotation information,
our task is to decode the category-level rotation feature. To achieve this, we utilize two decoders to extract the rotation information in a decoupled fashion. 
The two decoders decode the rotation information into two perpendicular vectors under corresponding rotation. These two vectors can represent rotation information completely (shown in Figure \ref{fig:axis_0}).

Since the two vectors are orthogonal, the decoded rotation information related to them is independent; we can use one of them to recover part rotation information of the object. For example, in Figure \ref{fig:cate_show}, we use the green vector axis to recover the pose. We can see that the green boxes and blue boxes are aligned well in the recovered axis.

Each decoder only needs to extract the orientation information along corresponding vector which is easier than the estimation of the complete rotation. The loss function is based on cosine similarity that defined as
\begin{equation}
\mathcal{L}_{rot} =  \frac{\left \langle \hat{\textbf{v}}_1,\textbf{v}_1 \right \rangle}{\| \hat{\textbf{v}}_1  \| \| \textbf{v}_1  \|} + \lambda_r\frac{\left \langle \hat{\textbf{v}}_2,\textbf{v}_2 \right \rangle}{\| \hat{\textbf{v}}_2  \| \| \textbf{v}_2  \|},
\end{equation}
where $\hat{\textbf{v}}_1$ and $\hat{\textbf{v}}_2$ are the predicted vectors. $\textbf{v}_1$ and $\textbf{v}_2$ are the ground truth, and $\lambda_r$ is the balance parameter. 

The balance parameter $\lambda_r$ makes our network easy to handle circular symmetry object such as bottle, and for such circular symmetry object, the red vector is not necessary (shown in Figure \ref{fig:axis_0}).
Without loss of generality, we assume that the green vector is along the symmetry axis; then, we set $\lambda_r$ as zero to handle the circular symmetry objects. For other types of symmetric objects, we can employ the rotation mapping function used in \cite{pitteri2019object, tian2020shapeprior} to map the relevant rotation matrices to a unique one. 

{Please note that our decoupled rotation is different to the rotation representation proposed in \cite{zhou2019continuity}. They took the first two columns from a rotation matrix as the new representation, which has no geometric meaning. 
In contrast, our representation is defined based on the shape of the target object, and our representation can avoid the discontinuity issue mentioned in \cite{zhou2019continuity, pitteri2019object}.}

\subsection{Residual Prediction Network}
\label{sec:res_net}
As both translation and object size are related to points coordinates, inspired by \cite{Qi_2018_CVPR,Chen_2020_CVPR}, we train a tiny PointNet \cite{Qi_2017_CVPR} that takes segmented point cloud as input. More concretely, the PointNet performs two related tasks: 1) estimating the residual between the translation ground truth and the mean value of the segmented point cloud; 2) estimating the residual between object size and the mean category size. 

For size residual, we pre-calculate the mean size $[\overline{x}, \overline{y}, \overline{z}]^T$ of each category by
\begin{equation}
\begin{bmatrix}
\overline{x}\\ 
\overline{y}\\ 
\overline{z}
\end{bmatrix} = \frac{1}{N}\sum_{i=1}^{N}[{x_i},{y_i},{z_i}]^T,
\end{equation}
where $N$ is the amount of the object in that category. Then for object $o$ in that category the ground truth $[\delta_x^o,
\delta_y^o,
\delta_z^o]^T$ of the size residual estimation is calculated as
\begin{equation}
[\delta_x^o,
\delta_y^o,
\delta_z^o]^T = [x_o,
y_o, 
z_o]^T - [
\overline{x},
\overline{y},
\overline{z}]^T.
\end{equation}

We use mean square error (MSE) loss to predict both the translation and size residual. The total loss function $\mathcal{L}_{res}$ is defined as:
\begin{equation}
\mathcal{L}_{res} = \mathcal{L}_{tra}+\mathcal{L}_{size},
\end{equation}
where $\mathcal{L}_{tra}$ and $\mathcal{L}_{size}$ are sub-loss for translation residual and size residual, respectively. 

\subsection{3D Deformation Mechanism}
One major problem in category-level 6D pose estimation is the intra-class shape variation. The existing methods employed two large synthetic datasets, i.e. CAMERA \cite{wang2019nocs} and 3D model dataset \cite{chang2015shapenet} to learn this variation. 
However, this strategy not only needs extra hardware resources to store these big synthetic datasets but also increases the (pre-)training time.

\begin{figure}[t!]
\begin{center}
\includegraphics[width=0.99\linewidth]{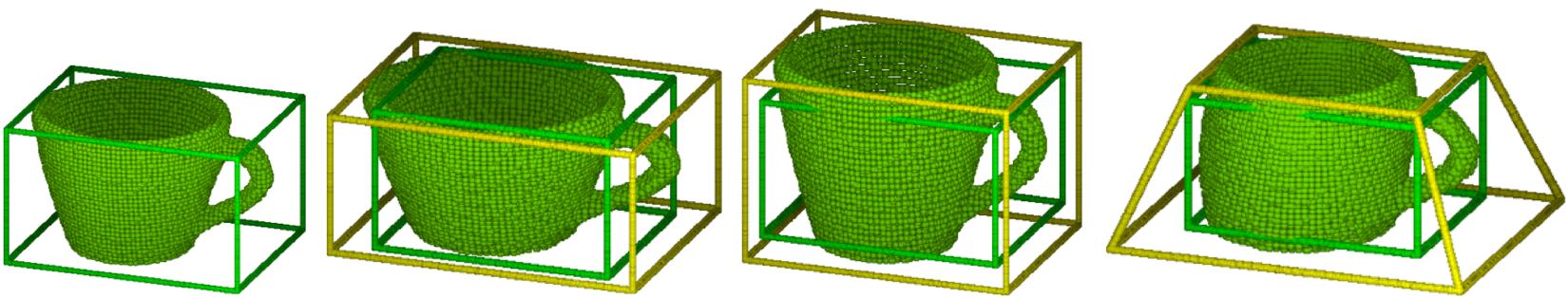}

\end{center}
\hspace{-10pt}
\vspace{-10pt}
\caption{\textbf{3D deformed examples}. The new training examples can be generated by enlarging, shrinking, or changing the area of some surfaces of the box-cages. 
The left one is the original point could with original 3D box-cage, i.e. 3D bounding box. The right three ones are the deformed point clouds with deformed box-cages (shown in yellow color). The green boxes are the original 3D bounding boxes before deformation.}
\label{fig:3d_defor}
\end{figure}

To alleviate the shape variation issue, based on the fact that the shapes of most objects in the same category are similar \cite{vlach2016we} (shown in Figure \ref{fig:color_var}), we propose an online box-cage based 3D deformation mechanism for training data augmentation. We pre-define a box-cage for each rigid object (shown in Figure \ref{fig:3d_defor}). Each point is assigned to its nearest surface of the cage; when we deform the surface, the corresponding points move as well.

Though box-cage can be designed more refined, in experiments, we find that with a simple box cage, i.e. 3D bounding box of the object, the generalization ability of the proposed method is considerably improved (Table \ref{tab:abla}). 
Different to \cite{yifan2020neural}, we do not need the extra training process to obtain the box-cage of the object, and we do not need target shape to learn the deformation operation either. Our mechanism is totally online, which saves training time and storage space.

To make the deformation operation easier, we first transfer the points to the canonical coordinate system and then perform 3D deformation. Finally we transform them to global scene:
\begin{equation}
    \{\mathcal{P}_1,\mathcal{P}_2,\cdots ,\mathcal{P}_n\}=R(\mathbb{F}_{3D}(R^T(\mathcal{P}-T)))+T,
\end{equation}
where $\mathcal{P}$ is the points generated after the 2D detection step. $R$, $T$ are the pose ground truth. $\{\mathcal{P}_1,\mathcal{P}_2,\cdots ,\mathcal{P}_n\}$ are the new generated training examples. $\mathbb{F}_{3D}$ is 3D deformation which includes cage enlarging, shrinking, changing the area of some surfaces.

\section{Experiments}

\subsection{Datasets}
\noindent \textbf{NOCS-REAL} \cite{wang2019nocs} is the first real-world dataset for category-level 6D object pose estimation. The training set has 4300 real images of 7 scenes with 6 categories. For each category, there are 3 unique instances. In the testing set, there are 2750 real images spread in 6 scenes of the same 6 categories as the training set. In each test scene, there are about 5 objects which makes the dataset clutter and challenging.

\noindent \textbf{LINEMOD} \cite{Hinterstoisser2013} is a widely used instance-level 6D object pose estimation dataset which consists of 13 different objects with significant shape variation.

We use the automatic point-wise labeling techniques proposed in \cite{Chen_2020_WACV} to access the label of each point in both training sets.

\subsection{Implementation Details}
We use Pytorch \cite{paszke2017pytorch} to implement our pipeline. All experiments are deployed on a PC with i7-4930K 3.4GHz CPU and GTX 1080Ti GPU.

First, to locate the object in RGB images, we fine-tune the YOLOv3 pre-trained on COCO dataset \cite{lin2014microsoft} with the training dataset.
Then we jointly train the 3DGC autoencoder and residual estimation network. The total loss function is defined as
\begin{equation}
\mathcal{L}_{Shape} =  \lambda_{seg} \mathcal{L}_{seg}+\lambda_{rec} \mathcal{L}_{rec} +\lambda_{rot} \mathcal{L}_{rot} +\lambda_{res} \mathcal{L}_{res},
\end{equation}
where $\lambda$s are the balance parameters. We empirically set them as 0.001, 1, 0.001, and 1 to keep different loss values at the same magnitude. We use cross entropy for 3D segmentation loss function $\mathcal{L}_{seg}$.

We adopt Adam \cite{kingma2014adam} to optimize the FS-Net. The initial learning rate is 0.001, and we halve it every 10 epochs. The maximum epoch is 50.

\subsection{Evaluation Metrics}
For category-level pose estimation, we adopt the same metrics used in \cite{wang2019nocs, chen2020cass,tian2020shapeprior}:
\begin{itemize}
\item$IoU_{X}$ is Intersection-over-Union (IoU) accuracy for 3D object detection under different overlap thresholds. The overlap ratio larger than the threshold $X$ is accepted.

\item $n^{\circ}$ $m$ $\textbf{cm}$ represents pose estimation error of rotation and translation. The rotation error less than $n^{\circ}$ and the translation error less than $m$ $\textbf{cm}$ is accepted.
\end{itemize}

For instance-level pose estimation, we compare the performance of FS-Net with other state-of-the-art instance-level methods using the ADD-(S) metric \cite{Hinterstoisser2013}.

\subsection{Ablation Studies}

\begin{table}[t]
\caption{\textbf{Ablation studies on NOCS-REAL dataset}. We use two different metrics to measure performance. `3DGC' means the 3D graph convolution. `OPR' means observed points reconstruction. `DR' represents the decoupled rotation mechanism. `DEF' denotes the online 3D deformation. In the last row, the values in the bracket are the performance for the reconstruction of the complete object model transformed by the corresponding pose. Please note, for the sake of ablation study, we provide the ground truth 2D bounding box for different methods.}
\label{tab:abla}
\vspace{2mm}
\centering
\resizebox{0.99\linewidth}{12mm}{
\begin{tabular}{|l|cccccc|}
\hline
Method & 3DGC& DEF &  OPR  & DR& $IoU_{50}$ & 10$^{\circ}$10 \textbf{cm}\\
\hline\hline
G2L \cite{Chen_2020_CVPR} & $\times$ &$\checkmark$& $\times$ & $\times$& 94.65\% & 31.0\% \\
G2L+DR & $\times$ &$\checkmark$& $\times$ & $\checkmark$& 96.21\% & 47.81\% \\
Med1 &$\checkmark$ &$\checkmark$ &  $\times$ &  $\times$& 97.98\%&46.4\% \\
Med2 & $\checkmark$&$\checkmark$  & $\checkmark$ & $\times$ &95.61\% &46.8\%\\
Med3 & $\checkmark$ &$\checkmark$ &  $\times$ & $\checkmark$& 97.34\%&61.1\%\\
Med4 & $\checkmark$ &$\times$ & $\checkmark$ & $\checkmark$&97.30\% & 58.2\%\\
Med5 & $\checkmark$ &$\checkmark$ & $\checkmark$ & $\checkmark$&98.04\% (94.44\%) &65.9\% (58.0\%)\\
\hline
\end{tabular}
}
\end{table}

We use the G2L-Net \cite{Chen_2020_CVPR} as the baseline method which extracted the latent feature for rotation estimation via point-wise orientated vector regression, and the ground truth of rotation is the eight corners of 3D bounding box with corresponding rotation. The loss function for rotation estimation is the mean square error between predicted 3D coordinates and ground truth.
Compared to baseline, our proposed work has three novelties: a) view-based 3DGC autoencoder for observed point cloud reconstruction; b) rotation decoupled mechanism; c) online 3D deformation mechanism.

In Table \ref{tab:abla}, we report the experimental results of three novelties on the NOCS-REAL dataset. Comparing Med3 and Med5, we find that reconstruction of the observed point cloud can learn better pose feature. The performance of Med2(Med1, G2L) and Med5(Med3, G2L+DR) shows that the proposed decoupled rotation mechanism can effectively extract the rotation information. The results of Med4 and Med5 demonstrate the effectiveness of the 3D deformation mechanism, which increases the pose accuracy by $7.7\%$ in terms of 10$^{\circ}$10 \textbf{cm} metric.
We also compare the different reconstruction choices: the reconstruction of observed points and the complete object model with corresponding rotation. From the last row of Table \ref{tab:abla}, we can see that the observed points reconstruction can learn better rotation feature. Overall, Table \ref{tab:abla} shows that the proposed novelties can improve the accuracy significantly.

\subsection{Generalization Performance}
\begin{figure}[htp!]
\begin{center}
\includegraphics[width=0.7\linewidth]{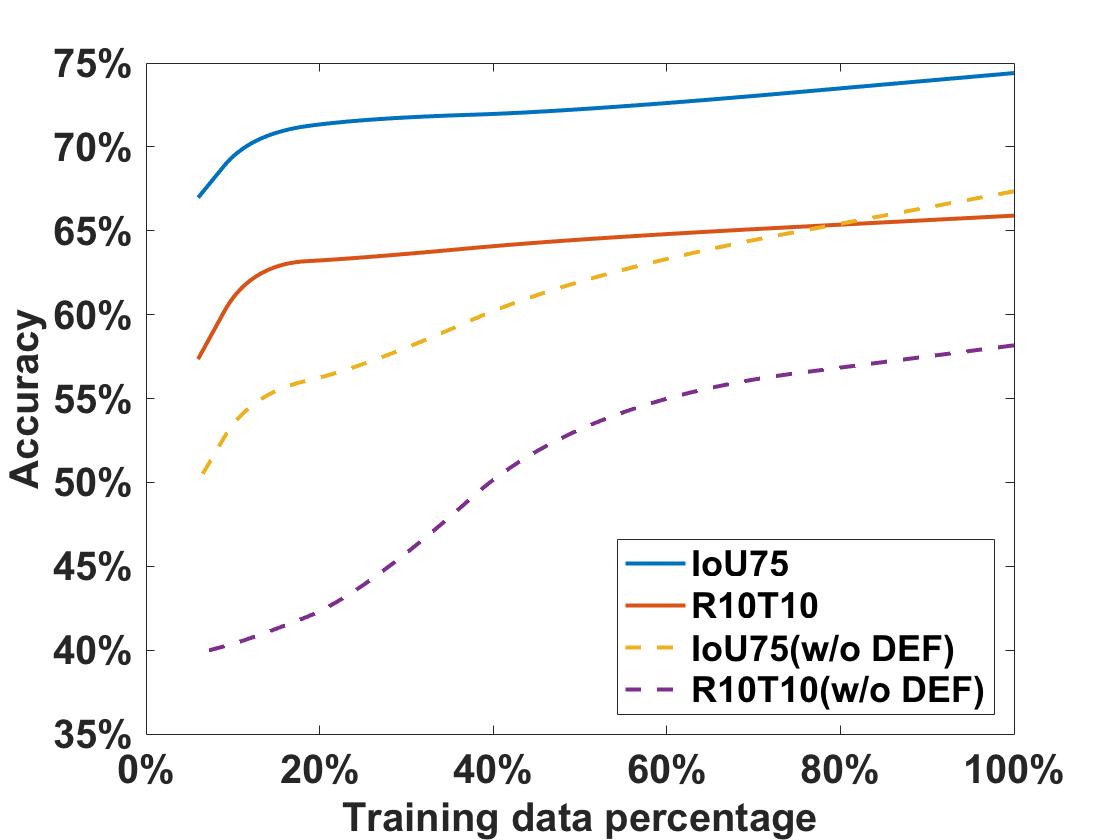}

\end{center}
   \caption{\textbf{Generalization performance.} With the given 2D bounding box and a randomly chosen 3D sphere center, we show how the training set size affects the pose estimation performance. `w/o DEF' means no 3D deformation mechanism is adopted during training.}
\label{fig:per}
\end{figure}

NOCS-REAL dataset provides 4.3k real images that covers various poses of different objects in different categories for training. That means the category-level pose information is rich in the training set. Thanks to the effectively pose feature extraction, FS-Net achieves state-of-the-art performance even with part of the real-world training data. We randomly choose different percentages of the training set to train FS-Net and test it on the whole testing set. Figure \ref{fig:per} shows that: 1) FS-Net is robust to the size of the training dataset, and has good category-level feature extraction ability. Even with $20\%$ of the training dataset, the FS-Net can still achieve state-of-the-art performance; 2) the 3D deformation mechanism significantly improves the robustness and performance of FS-Net.

\subsection{Evaluation of Reconstruction}
Point cloud reconstruction has a close relationship with pose estimation performance. We compute the Chamfer Distance of the reconstructed point cloud with the ground truth point cloud and compared it with other reconstruction types used by other methods. From Table \ref{tab:recon}, we can see that the average reconstruction error of our method is 0.86, which is $72.9\%$ and $18.9\%$ lower than that of Shape-Prior \cite{tian2020shapeprior} and CASS \cite{chen2020cass}, respectively. It shows that our method achieves better pose estimation results via a simpler reconstruction task, i.e. observed points reconstruction rather than complete object model reconstruction.

\subsection{Comparison with State-of-the-Arts}

\subsubsection{Category-Level Pose Estimation}
\begin{figure*}[htp!]
\begin{center}
\includegraphics[width=0.9\linewidth]{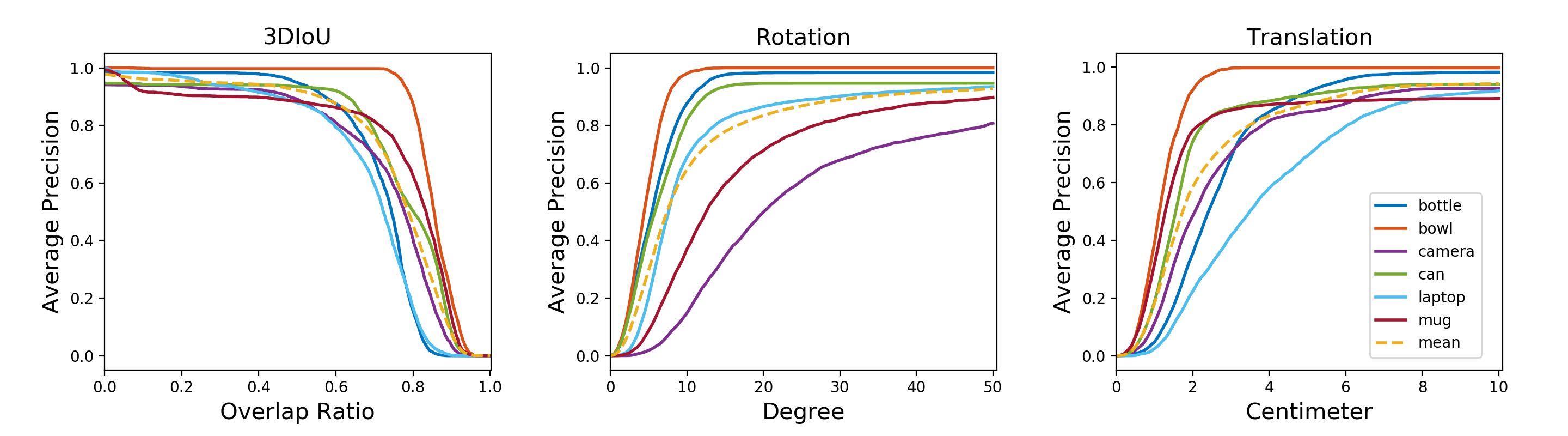}
\caption{\textbf{Result on NOCS-REAL}. The average precision of different thresholds tested on NOCS-REAL dataset with 3D IoU, rotation, and translation error.}
\label{fig:iourt}
\end{center}
\vspace{-10pt}
\hspace{-10pt}
\end{figure*}

\begin{figure*}[t!]
\begin{center}
\resizebox{0.99\textwidth}{35mm}{
\begin{tabular}{cccccc}
{{\includegraphics[width=0.15\linewidth]{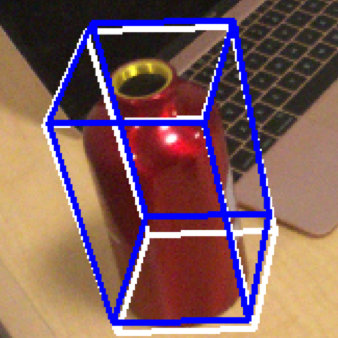}}}&
{{\includegraphics[width=0.15\linewidth]{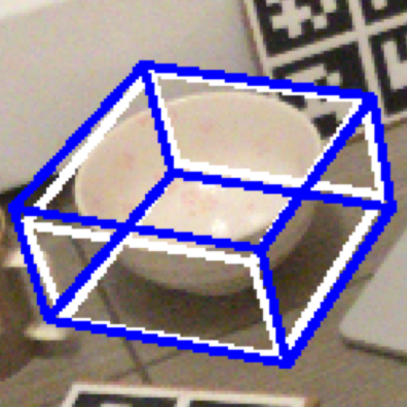}}}&
{{\includegraphics[width=0.15\linewidth]{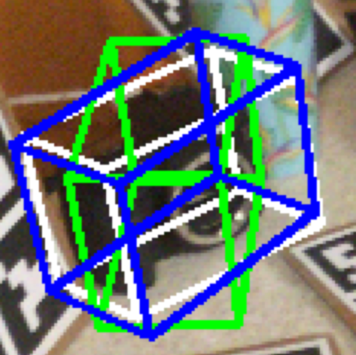}}}&
{{\includegraphics[width=0.15\linewidth]{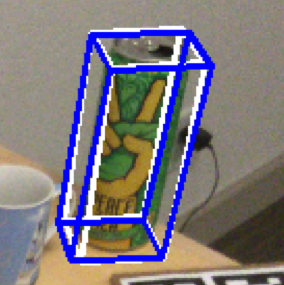}}}&
{{\includegraphics[width=0.15\linewidth]{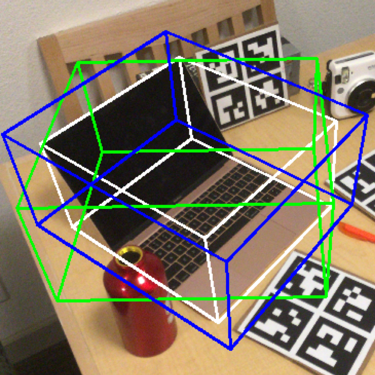}}}&
{{\includegraphics[width=0.15\linewidth]{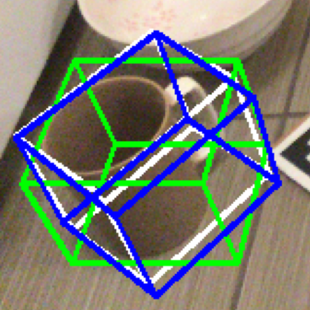}}}\\
{{\includegraphics[width=0.15\linewidth]{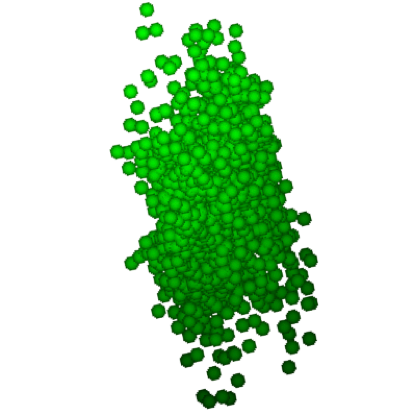}}}&
{{\includegraphics[width=0.15\linewidth]{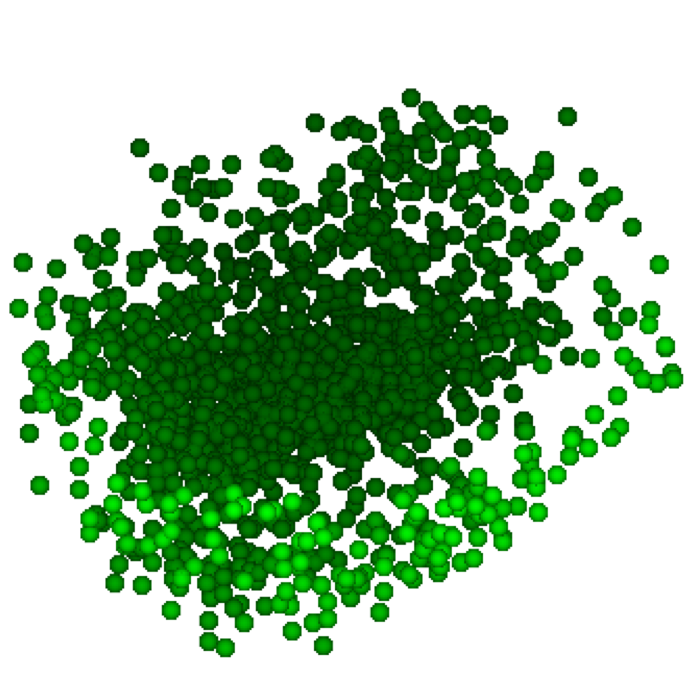}}}&
{{\includegraphics[width=0.15\linewidth]{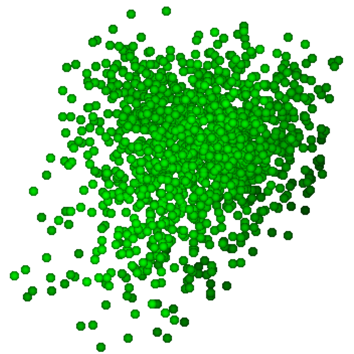}}}&
{{\includegraphics[width=0.15\linewidth]{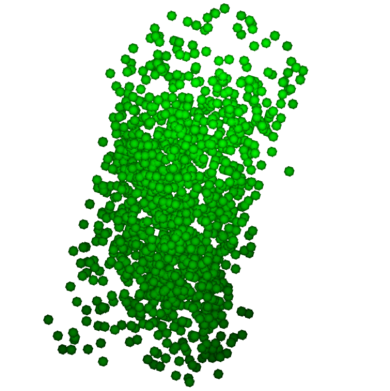}}}&
{{\includegraphics[width=0.15\linewidth]{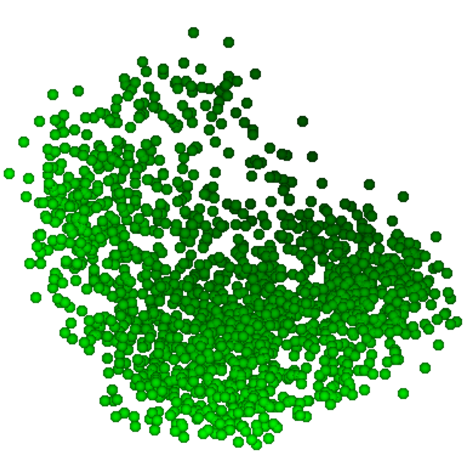}}}&
{{\includegraphics[width=0.15\linewidth]{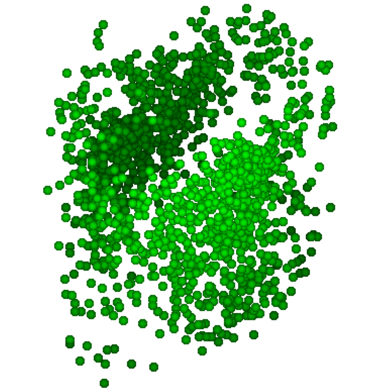}}}\\
{{\includegraphics[width=0.15\linewidth]{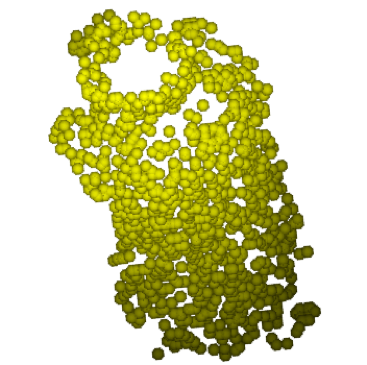}}}&
{{\includegraphics[width=0.15\linewidth]{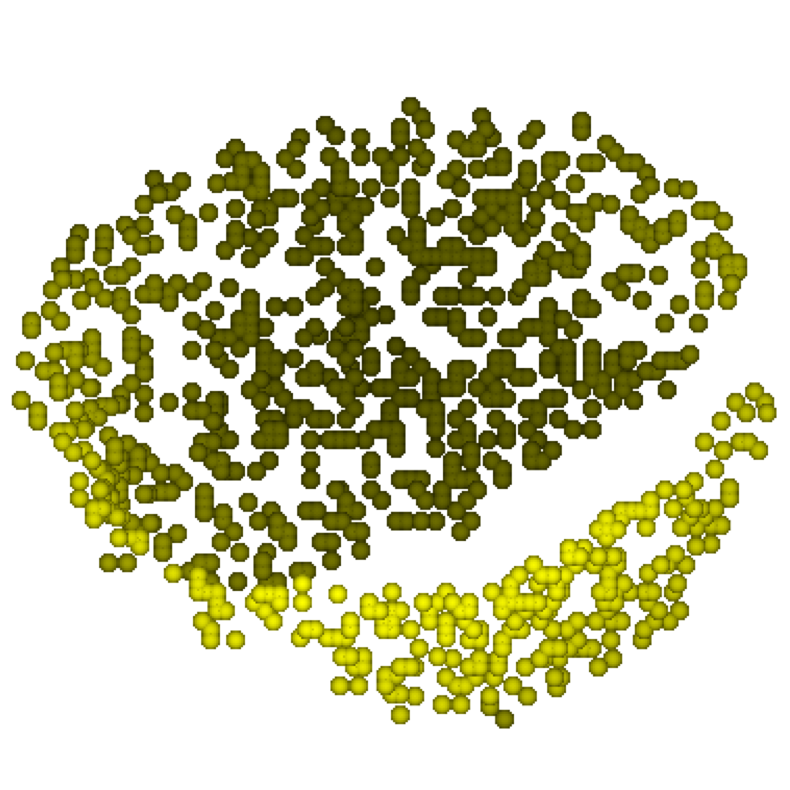}}}&
{{\includegraphics[width=0.15\linewidth]{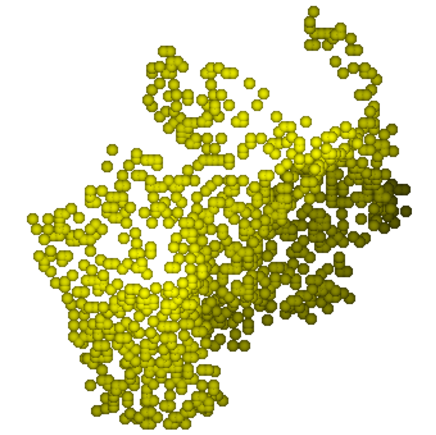}}}&
{{\includegraphics[width=0.15\linewidth]{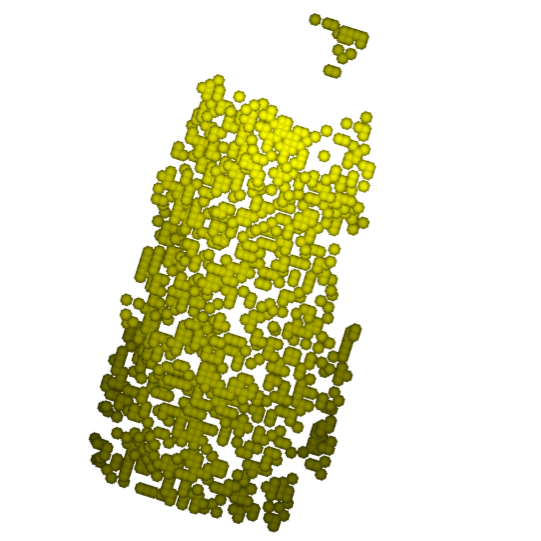}}}&
{{\includegraphics[width=0.15\linewidth]{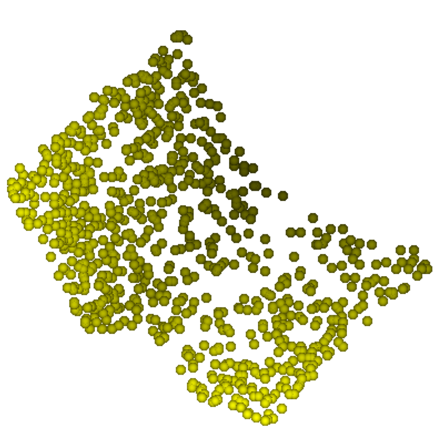}}}&
{{\includegraphics[width=0.15\linewidth]{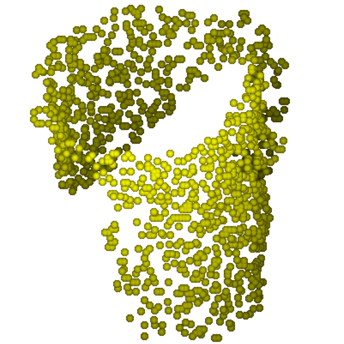}}}\\
\end{tabular}}
\end{center}
\vspace{-10pt}
\hspace{-10pt}
\caption{\textbf{Qualitative results on NOCS-REAL dataset}. The first row is the pose and size estimation results. White 3D bounding boxes denote ground truth. Blue boxes are the poses recovered from two estimated rotation vectors. The green boxes are the poses recovered from one estimated rotation vector. Our results match ground truth well in both pose and size. The second row is the reconstructed observed points under corresponding poses, although the reconstructed points are not perfectly in line with the target points, the basic orientation information is kept. The third row is the ground truth of the observed points transformed from the observed depth map.}
\label{fig:cate_show}
\end{figure*}

We compare FS-Net with NOCS \cite{wang2019nocs}, CASS \cite{chen2020cass}, Shape-Prior \cite{tian2020shapeprior}, and 6D-PACK \cite{wang20206dpack} on NOCS-REAL dataset in Table \ref{tab:catepose}. We can see that our proposed method outperforms the other state-of-the-art methods on both accuracy and speed. Specifically, on 3D detection metric $IOU_{50}$, our FS-Net outperforms the previous best method, NOCS, by $11.7\%$ and the running speed is 4 times faster. In terms of 6D pose metric 5$^{\circ}$5\textbf{cm} and 10$^{\circ}$10 \textbf{cm}, FS-Net outperforms the CASS by the margins of $4.7\%$ and $6.3\%$, respectively. FS-Net even outperforms 6D-PACK under 3D detection metric $IOU_{50}$, which is a 6D tracker and needs an initial 6D pose and object size to start. See Figure \ref{fig:iourt} for more quantitative details. The qualitative results are shown in Figure \ref{fig:cate_show}. Please note, we only use real-world data (NOCS-REAL) to train our pose estimation part. Other methods use both synthetic dataset (CAMERA) \cite{wang2019nocs} and real-world data for training. The number of training examples in CAMERA is 275K, which is more than 60 times that of NOCS-REAL (4.3K). It shows that FS-Net can efficiently extract the category-level pose feature with fewer data.

\begin{table}[t]
\caption{\textbf{Reconstruction type comparison.} The comparison is on the NOCS-REAL dataset with the Chamfer Distance metric ($\times 10^{-3}$). `Complete' means the reconstruction of the complete 3D model. `Observed' denotes only the reconstruction of the observed points. }
\label{tab:recon}
\vspace{1mm}
\centering
\begin{tabular}{|l|ccc|}
\hline
Methods &  CASS \cite{chen2020cass} & Shape-Prior \cite{tian2020shapeprior}& Ours\\
\hline\hline
  & Complete & Complete & Observed\\
Bottle &0.75 & 3.44& 1.2 \\
 Bowl &   0.38 &1.21 &  0.39 \\
 Camera &0.77&   8.89 &\textbf{0.44} \\
  Can &  0.42 &  1.56 &0.62\\
  Laptop & 3.73 &2.91 & \textbf{2.23} \\
  Mug &0.32& 1.02 & \textbf{0.29} \\
  Average&1.06 &3.17&\textbf{0.86} \\
\hline
\end{tabular}
\end{table}

\begin{table}[htb]
\caption{\textbf{Instance-level comparison on LINEMOD dataset.} Our method achieves a comparable performance with the state-of-the-art in both speed and accuracy.}
\label{tab:linmod}
\vspace{1mm}
\centering
\begin{tabular}{|l|ccc|}
\hline
Method & Input & ADD-(S) & Speed(FPS)\\
\hline\hline
PVNet \cite{peng2018pvnet}& RGB &  86.3\% & 25 \\
CDPN \cite{li2019cdpn}& RGB &  89.9\% & 33 \\
DPOD \cite{zakharov2019dpod}& RGB &  95.2\% & 33 \\
G2L-Net \cite{Chen_2020_CVPR} & RGBD &  98.7\% & 23 \\
Densefusion\cite{wang2019densefusion} & RGBD &  94.3\% & 16  \\
PVN3D \cite{he2020pvn3d} &RGBD&  99.4\% &5 \\
Ours & RGBD&  {97.6\%} & 20 \\
\hline
\end{tabular}
\end{table}

\begin{table*}[htb!]
\centering
\caption{\textbf{Category-level performance on NOCS-REAL dataset with different metrics.} We summarize the pose estimation results reported in the origin papers on the NOCS-REAL dataset. `-' means no results are reported under this metric. The values in the bracket are the performance for synthetic NOCS dataset.}
\label{tab:catepose}
\vspace{1mm}
\begin{tabular}{|l|ccccccc|}
\hline
Method &$IoU_{25}$ & $IoU_{50}$ & $IoU_{75}$ & 5$^{\circ}$5\textbf{cm}& 10$^{\circ}$5 \textbf{cm}& 10$^{\circ}$10 \textbf{cm}& Speed(FPS)\\
\hline\hline
NOCS \cite{wang2019nocs} & 84.9\%&  80.5\% & 30.1\%(69.5\%) &  9.5 \%(40.9\%) & 26.7\%& 26.7\%&  5 \\
CASS \cite{chen2020cass} & 84.2\%&  77.7\% & - &  23.5 \% & 58.0\%& 58.3\%&   - \\
Shape-Prior \cite{tian2020shapeprior} &83.4\%&  77.3\% & 53.2\%(83.1\%) &  21.4\%(59.0\%) & 54.1\%& -&  4\\
6-PACK \cite{wang20206dpack} & 94.2\%&  - & - &  \textbf{33.3 \%} & -& -&  10\\
Ours & \textbf{95.1\%}&  \textbf{92.2\%} & \textbf{63.5\%}(85.17\%)&  {28.2 \%}(62.01\%) & \textbf{60.8\%}& \textbf{64.6\%}&  \textbf{20}\\
\hline
\end{tabular}
\end{table*}

\subsubsection{Instance-Level Pose Estimation}
We compare the instance-level pose estimation results of FS-Net on the LINEMOD dataset with other state-of-the-arts instance-level methods. From Table \ref{tab:linmod}, we can see that FS-Net achieves comparable results on both accuracy and speed. It shows that our method can effectively extract both category-level and instance-level pose features.

\subsection{Running Time}
Given a 640$ \times$ 480 RGB-D image, our method runs at 20 FPS with Intel i7-4930K CPU and 1080Ti GPU, which is 2 times faster than the previous fastest method 6D-PACK \cite{wang20206dpack}. Specifically, the 2D detection takes about 10ms to proceed. The pose and size estimation takes about 40ms.


\section{Conclusion}
In this paper, we propose a fast category-level pose estimation method that runs at 20 FPS which is fast enough for real-time applications. The proposed method first extracts the latent feature by the observed points reconstruction with a shape-based 3DGC autoencoder. Then the category-level orientation feature is decoded by the effective decoupled rotation mechanism. Finally, for translation and object size estimation, we use the residual network to estimate them based on residuals estimation. In addition, to increase the generalization ability of FS-Net and save the hardware source, we design an online 3D deformation mechanism for training set augmentation. Extensive experimental results demonstrate that FS-Net is less data-dependent, and can achieve state-of-the-art performance on category- and instance-level pose estimation in both accuracy and speed. Please note, our 3D deformation mechanism and decoupled rotation scheme are model-free, which can be applied to other pose estimation methods to boost the performance. 

Although FS-Net achieves state-of-the-art performance, it relies on a robust 2D detector to detect the region of interest. In future work, we plan to adopt 3D object detection techniques to directly detect the objects from point clouds.


\section{Appendix}

This section provides more details about our FS-Net. Section \ref{sec:3df} describes the details of the 3D deformation mechanism and deformed examples. Section \ref{sec:exp} provides more quantitative results of the FS-Net on NOCS-REAL \cite{wang2019nocs} dataset and comparison with state-of-the-art method. Section \ref{sec:DR} demonstrates that the proposed vectors-based rotation representation can be easily extended to handle other symmetric types.

\subsection{3D Deformation Mechanism}
\label{sec:3df} 
\begin{figure}[t!]
\centering
\begin{tabular}{c}
{{\includegraphics[width=0.9\linewidth]{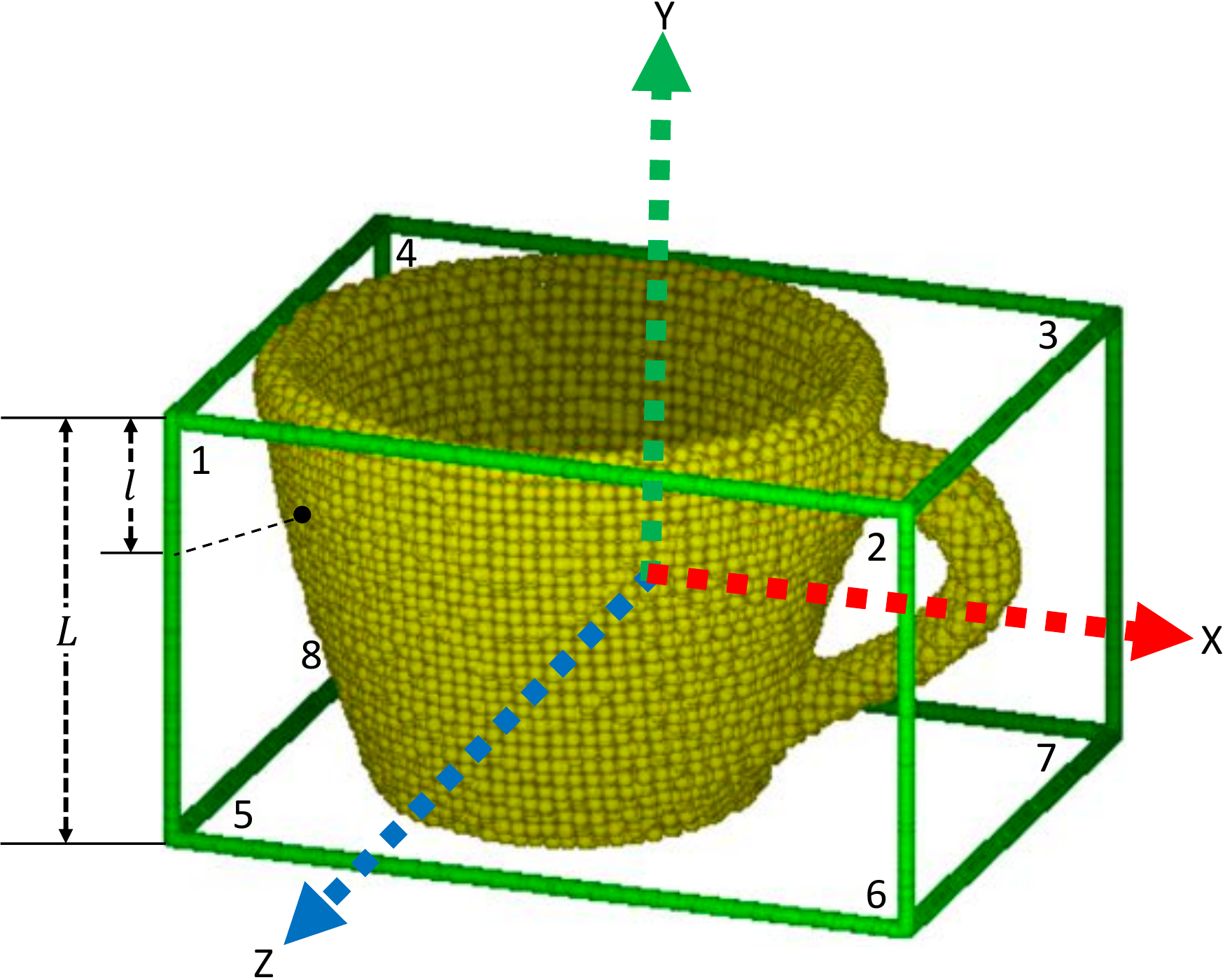}}}
\end{tabular}
\caption{\textbf{3D object model}. We assume that the center of 3D bounding box is the origin point of the coordinate. The surface is represented by its four corners. For example, the top surface is represented by $S_{1,2,3,4}$.}
\label{fig:3D_axis}
\end{figure}

As stated in Section 3.5 of the paper, the 3D deformation mechanism is box-cage based and the deformations are applied in a canonical space. In the canonical coordinate system, every box edge is parallel to an axis (shown in Figure \ref{fig:3D_axis}). This property makes the 3D deformation calculation easier. For example, when we need to elongate/shrink the mug along $Y$ axis by $n$ times. We enlarge the distance between surface $S_{1,2,3,4}$ and surface $S_{5,6,7,8}$ by $n$ times. Since these two surfaces are parallel to the $XZ$-plane, the $x$ and $z$ coordinates are unchanged. 
Then points coordinates are changed from $[\textbf{x}, \textbf{y}, \textbf{z}]$ to $[\textbf{x}, n\textbf{y}, \textbf{z}]$. The calculations are similar when we need to elongate/shrink the mug along $X$ or $Z$ axis by $n$ times:
\begin{align}
[\textbf{x}, n \textbf{y}, \textbf{z}] = \mathbb{F}_{x} ([\textbf{x}, \textbf{y}, \textbf{z}]),\\
[n \textbf{x}, \textbf{y}, \textbf{z}] = \mathbb{F}_{y}( [\textbf{x}, \textbf{y}, \textbf{z}]),\\
[\textbf{x}, \textbf{y}, n \textbf{z}] = \mathbb{F}_{z} ([\textbf{x}, \textbf{y}, \textbf{z}]),
\end{align}
where $\mathbb{F}_{x,y,z}$ is the elongate/shrink operation along corresponding axis.

\begin{figure*}[t!]
\centering
\begin{tabular}{c}
{{\includegraphics[width=0.9\linewidth]{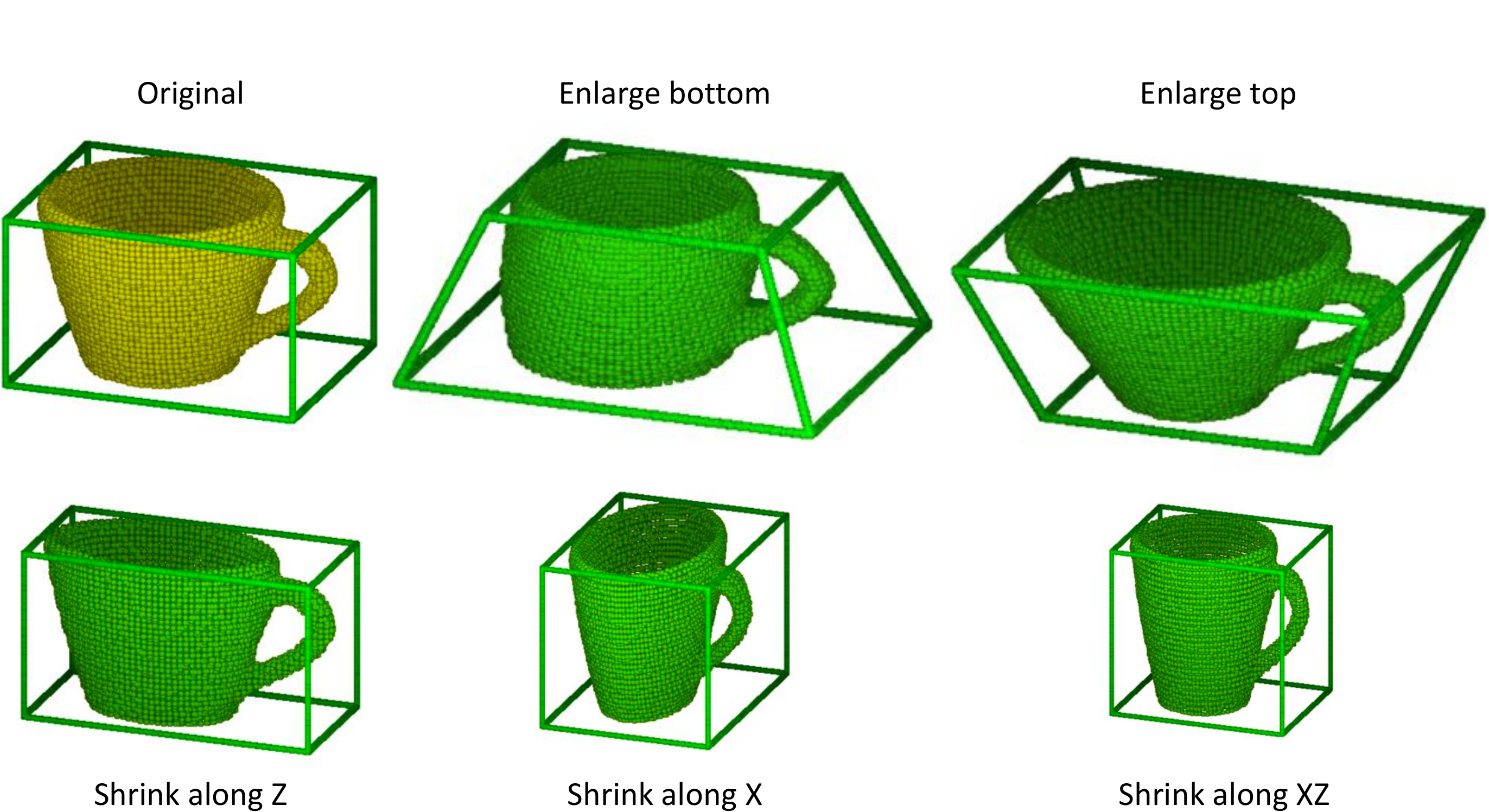}}}
\end{tabular}
\caption{\textbf{Examples of different deformations.} We assume that the $XYZ$ axis are the same as Figure \ref{fig:3D_axis}. The upper right corner is the original point cloud with corresponding box-cage. The rest are the deformed box-cages and point clouds. The deformation operations are described on the top or bottom of the pictures.}
\label{fig:3D_def}
\end{figure*}

Further, if the object is the mug or bowl, we may need to change the top or bottom size to generate new shapes (shown in Figure \ref{fig:3D_def}).  In this case, assuming we enlarge the bottom along $X$ axis by $n$ times, then from bottom to top, the coordinates are changed as:
\begin{equation}
\textbf{x}_{new} = (1+(n-1)\frac{l}{L})\textbf{x},
\end{equation}
where $l$ is the distance from a point to the top surface, i.e. $S_{1,2,3,4}$ in Figure \ref{fig:3D_axis}. $L$ is the height of the object. Please note, all the edges are keep straight while deformation.

\subsection{Experimental Results}
\label{sec:exp}

\subsubsection{Detailed Results}
\begin{figure*}[t!]
\centering
\begin{tabular}{c}
{{\includegraphics[width=0.9\linewidth]{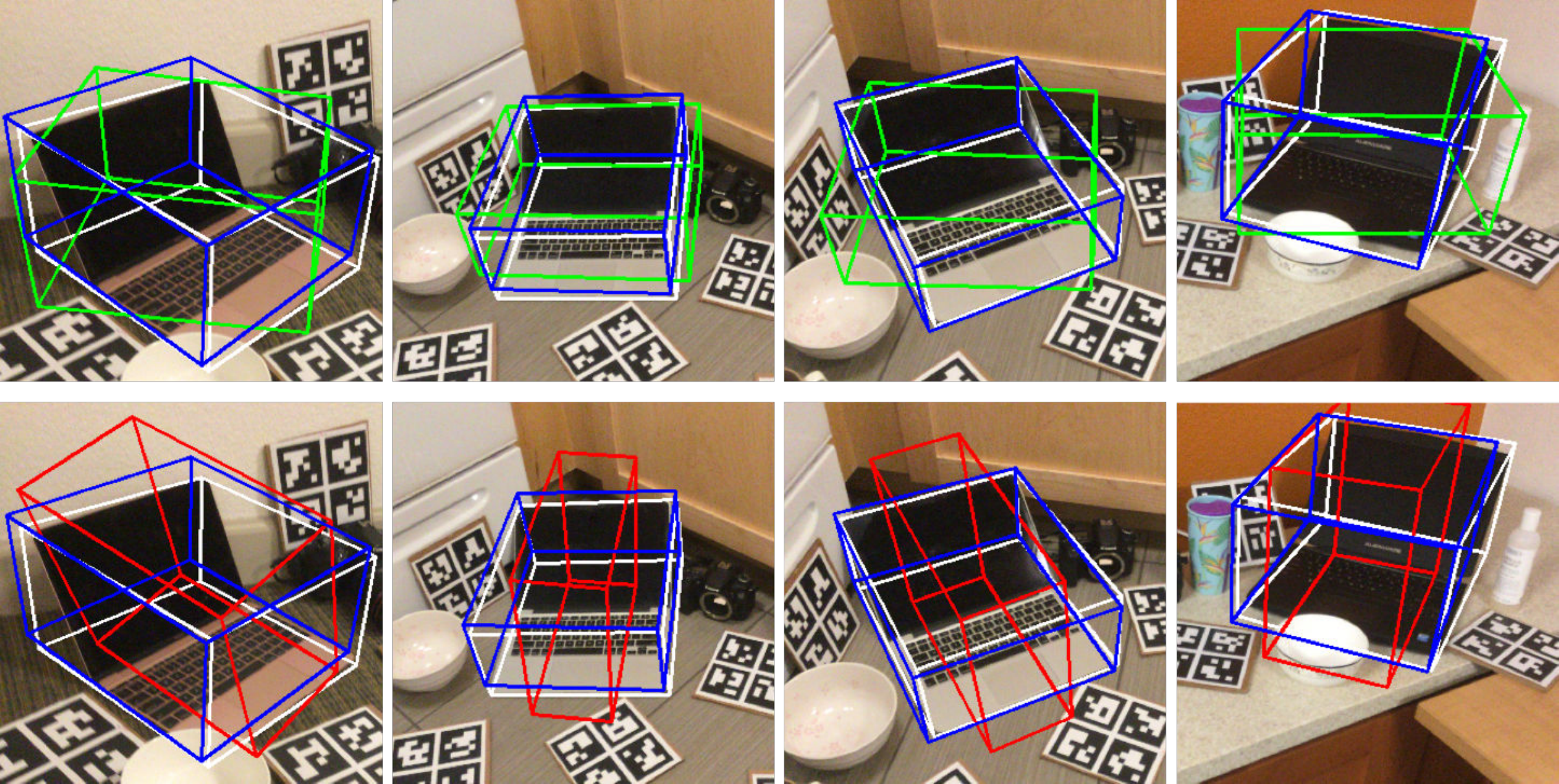}}}
\end{tabular}
\caption{\textbf{Rotation recovered by different vectors.}  The white boxes are the ground truth. Blue boxes are the rotation recovered by two estimated vectors. The green and red boxes are the rotation recovered by estimated green vector and estimated red vector (see Figure 4 in the paper), respectively. For better illustration, we use ground truth object size to calculate the final 3D bounding box.}
\label{fig:rot12}
\end{figure*}

We report the  specific category pose estimation results under different metrics in Table \ref{tab:object_w}. We also provide the rotation recovered by one/two vectors in Figure \ref{fig:rot12}.  We can see that the bounding boxes are well aligned in the recovered vector direction.

\begin{table}[htb]
\caption{\textbf{Category-Level results.} Object-wise experiments with different metrics.}
\label{tab:object_w}
\vspace{1mm}
\centering
\begin{tabular}{|l|cccc|}
\hline
Category &$IoU_{75}$&5$^{\circ}$5 \textbf{cm} &10$^{\circ}$5 \textbf{cm} & 10$^{\circ}$10 \textbf{cm}\\
\hline\hline
Bottle & 0.4710 &  0.4219 & 0.8134& 0.8755\\
Bowl & 0.9810 &  0.5916 &  0.9793&0.9793 \\
Camera & 0.5882 &  0.0176 & 0.1457&0.1480 \\
Can & 0.6334 &   0.4055 & 0.7820&0.8141 \\
Laptop & 0.3805 &  0.1659 & 0.5570& 0.6859 \\
Mug &0.7534&  0.0874 &0.3698& 0.3706\\
\hline
Average & 0.6345&   0.2816 &  0.6078&0.6455\\
\hline
\end{tabular}
\end{table}

\subsubsection{Comparison with  State-of-The-Art}
\begin{figure}[t]
\begin{tabular}{ccc}
{{\includegraphics[width=0.3\linewidth]{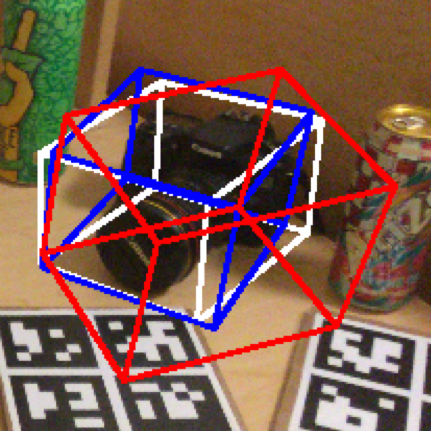}}}&
{{\includegraphics[width=0.3\linewidth]{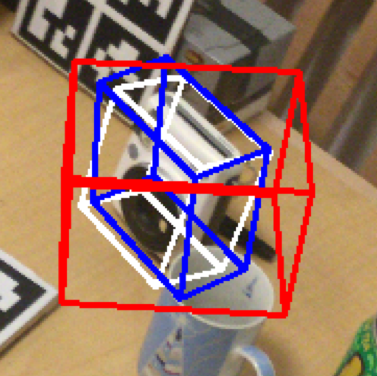}}}&
{{\includegraphics[width=0.3\linewidth]{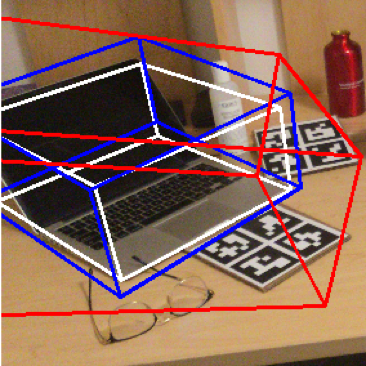}}}\\
{{\includegraphics[width=0.3\linewidth]{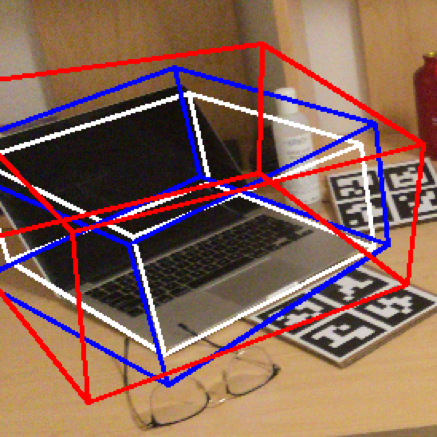}}}&
{{\includegraphics[width=0.3\linewidth]{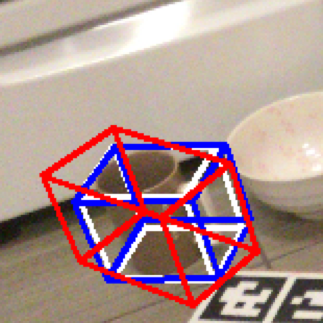}}}&
{{\includegraphics[width=0.3\linewidth]{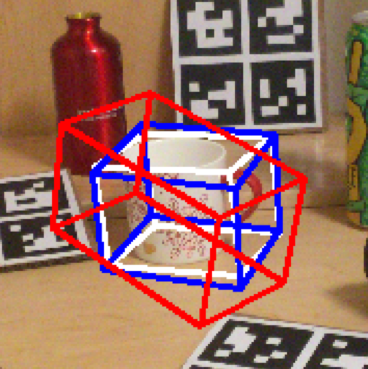}}}\\
\end{tabular}
\caption{\textbf{Qualitative comparison with Shape-Prior}. The white boxes are the ground truth. Blue boxes are our results. Red boxes are the poses predicted by Shape-Prior \cite{tian2020shapeprior}}
\label{fig:shp}
\end{figure}

We compare FS-Net with the state-of-the-art method Shape-Prior \cite{tian2020shapeprior}, which utilized point cloud for category-level 6D object pose estimation. Shape-Prior \cite{tian2020shapeprior} estimated the object size and 6D pose from dense-fusion feature \cite{wang2019densefusion}, while we estimate the pose from point cloud feature. Figure \ref{fig:shp} shows that our FS-Net is robust to color and shape variation, and can handle some failure cases of Shape-Prior. For Shape-Prior, we use the predicted results provided on their website: \url{https://github.com/mentian/object-deformnet}.

\subsection{Rotation Representation for Symmetry Object}
\label{sec:DR}
\begin{figure*}[ht]
\centering
	\includegraphics[width=0.99\linewidth]{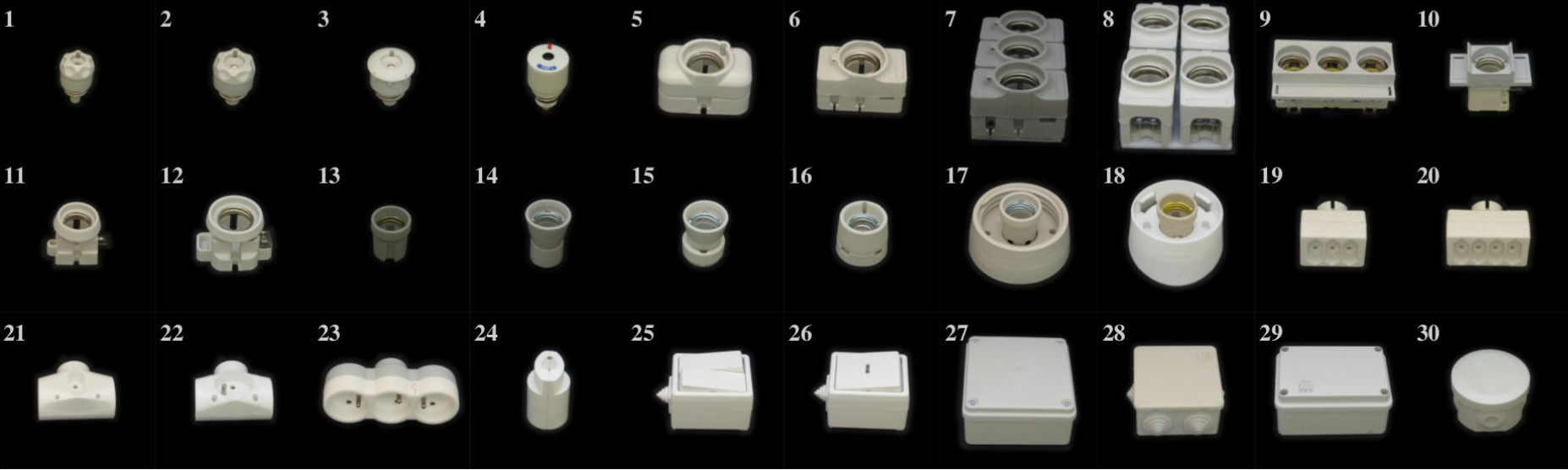}
	\caption{\textbf{Different symmetry types}. 30 industry-relevant objects in T-LESS dataset \cite{hodan2017t}. Object 1, 2, 3, 4 are circular symmetry, object 7, 8, 9, 10 have two symmetry axes, while object 27, 28 have four symmetry axes.}
\label{fig:symm}
\end{figure*}

The vector based rotation representation proposed in the paper can only handle the symmetry objects like bottle, however, in real-world the symmetric types are various (see Figure \ref{fig:symm}). 
In this section, we will show how to extend the vector based rotation representation for different symmetric types.
Our strategy is inspired by the rotation mapping operation proposed in \cite{pitteri2019object}. In the following, we will show how to find the rotation group (termed proper symmetries in \cite{pitteri2019object}) of a single rotation for common symmetric objects.

Our basic idea is list all the ambiguous rotations of a single rotation and choose the rotations that has the closest distance with the identity matrix:
\begin{equation}
\label{eq:best_r}
	R^* =\underset{R \in \mathcal{G}(R_i)}{\operatorname{arg min}} \mathcal{D}(R, R_I),
\end{equation}
where $\mathcal{D}(\cdot, \cdot)$  is the distance between two rotation matrix, $\mathcal{G}(R_i)$ is a group of rotation that can provide the same visual appearance of a given object as rotation $R_i$.
 Our goal is to find a rotation $R^*$ that can minimize the rotation distance.

For symmetric object like bottle, we can avoid the rotation ambiguity by only using the green vector to represent the rotation (see Figure \ref{fig:axis_0}), however, the case is non-trivial for other symmetric type. In the following, we describe how we find symmetry rotation group for different symmetric types


\subsubsection{Symmetry with two Axes}
\label{sec:sym_2}
For this kind symmetric objects, in canonical space, when we rotate the object around one axis 180$^{\circ}$, we can get the same appearance (see Figure for illustration). Assume that axis is $Z$ axis, for arbitrary rotation $R$, the appearance $\mathcal{A}$:
\begin{equation}
	\mathcal{A}^{R^{Z^{+}}_{180}\mathcal{O}} =\mathcal{A}^{\mathcal{O}},
\end{equation}
where $R^{Z^{+}}_{180}$ means rotation the object around $Z$ 180$^{\circ}$ in clockwise, $\mathcal{O}$ denotes the object. That means we can find the rotation group of each rotation by right multiplication operation $R^{Z^{+}}_{180}$. Then we use Equation \ref{eq:best_r} to find the representative rotation in the rotation group.
%

\subsubsection{Symmetry with $N$ Axes}
\label{sec:sym_n}
The idea can be easily extend to object with $N$ symmetries around a single axis $Z$.
For this kind of symmetric objects, when we rotate the object around axis $Z$ by $K \frac{360}{N}^{\circ} (K=1,2,\cdots,N)$ in canonical space, the appearance $\mathcal{A}$ of the object is unchanged:
\begin{equation}
	\mathcal{A}^{R^{Z^{+}}_{K \frac{360}{N}^{\circ} }\mathcal{O}} =\mathcal{A}^{\mathcal{O}}.
\end{equation}
Then, the symmetric rotation group $\mathcal{G}(R)$ of rotation $R$ is: $RR^{K \frac{360}{N}^{\circ} (K=0,1,2,\cdots,N)}$. We find the representative rotation in $\mathcal{G}(R)$ with Equation \ref{eq:best_r}.

\subsubsection{General Case}
Most symmetric types are included in the description of Section \ref{sec:sym_2} and \ref{sec:sym_n}. For any other symmetric object, the key idea here is to find the rotation operation that can produce the same appearance of the object. Then use Equation \ref{eq:best_r} to find the representative rotation.

\subsubsection{Decoupled Rotation Representation}
Given the representative rotation $R^*$ of ambiguous rotation, we generate its corresponding vector-based representation $\mathcal{V}$ by:
\begin{equation}
	\mathcal{V} = R^{*}[\textbf{v}_1,\textbf{v}_2],
\end{equation}
where $\textbf{v}_1$ is the vector along with the axis $Z$ mentioned in Section \ref{sec:sym_2} and \ref{sec:sym_n}, $\textbf{v}_2$ is the vectors orthogonal with $\textbf{v}_1$. 

{\small
\bibliographystyle{ieee_fullname}
\bibliography{egbib}
}

\end{document}